\documentclass[journal]{IEEEtran}
\usepackage{amsmath,amsfonts}
\usepackage{algorithm}
\usepackage{array}
\usepackage[caption=false,font=normalsize,labelfont=sf,textfont=sf]{subfig}
\usepackage{textcomp}
\usepackage{stfloats}
\usepackage{xcolor}
\usepackage{url}
\usepackage{verbatim}
\usepackage{graphicx}
\usepackage{svg}
\usepackage{cite}
\usepackage{booktabs} 
\usepackage{multirow} 
\usepackage{array} 
\usepackage{tabularx} 
\usepackage{algpseudocode} 
\usepackage{listings} 

\begin{document}

\title{Unsupervised Band Selection Using Fused HSI and LiDAR Attention Integrating With Autoencoder}

\author{Judy X~Yang,~\IEEEmembership{Student Member,~IEEE},
Jun~Zhou,~\IEEEmembership{Senior Member,~IEEE},
Jing~Wang,~\IEEEmembership{Senior Member,~IEEE},
Hui~Tian,~\IEEEmembership{Senior Member,~IEEE}, and~Alan Wee~Chung~Liew,~\IEEEmembership{Senior Member,~IEEE}}  



\maketitle

\begin{abstract}
Band selection in hyperspectral imaging (HSI) is critical for optimising data processing and enhancing analytical accuracy. Traditional approaches have predominantly concentrated on analysing spectral and pixel characteristics within individual bands independently. These approaches overlook the potential benefits of integrating multiple data sources, such as Light Detection and Ranging (LiDAR), and is further challenged by the limited availability of labeled data in HSI processing, which represents a significant obstacle.
To address these challenges, this paper introduces a novel unsupervised band selection framework that incorporates attention mechanisms and an Autoencoder for reconstruction-based band selection. Our methodology distinctively integrates HSI with LiDAR data through an attention score, using a convolutional Autoencoder to process the combined feature mask. This fusion effectively captures essential spatial and spectral features and reduces redundancy in hyperspectral datasets.
A comprehensive comparative analysis of our innovative fused band selection approach is performed against existing unsupervised band selection and fusion models. We used data sets such as Houston 2013, Trento, and MUUFLE for our experiments. The results demonstrate that our method achieves superior classification accuracy and significantly outperforms existing models. This enhancement in HSI band selection, facilitated by the incorporation of LiDAR features, underscores the considerable advantages of integrating features from different sources.

\end{abstract}


 \section{Introduction} 
Hyperspectral imaging technology captures intricate spectral responses of terrestrial objects across numerous narrow spectral bands~\cite{8314827,9645266}, finding widespread application in various real-world scenarios such as precision agriculture~\cite{6581194,sethy2022hyperspectral}, environmental monitoring~\cite{8314827,stuart2019hyperspectral}, and the prediction of earth disasters~\cite{9463743}. The primary goal of band selection in the field of hyperspectral imaging is to identify and isolate a concise subset of hyperspectral bands~\cite{8738051}. This selection process is essential for eliminating spectral redundancy and reducing computational demands while preserving essential spectral information pertinent to terrestrial objects, thereby ensuring the integrity and usefulness of the spectral data~\cite{10291338,9019632}.

The spectral richness captured by hyperspectral imaging (HSI) systems introduces a high-dimensionality and significant redundancy in HSI datasets, leading to computational and operational challenges. These include the Hughes phenomenon, characterised by a decline in classifier performance due to an excessively high feature-to-sample ratio~\cite{mashala2023systematic}, along with considerable computational burdens, and inefficiencies in data storage and transmission\cite{8738051}. To address these challenges, hyperspectral band selection serves as a vital preprocessing step. The purpose of this system is to extract the most informative spectral bands, removing redundant or unnecessary data. 

Extensive research in the field of hyperspectral imaging has identified various methodologies for band selection, dividing these approaches into five primary frameworks: ranking-based, search-based, clustering-based, deep learning-based and hybrid strategies~\cite{yu2021semisupervised,sawant2020unsupervised,wang2022hyperspectral}. Each framework offers a unique strategy to mitigate the challenges of high dimensionality and redundancy inherent in HSI data, highlighting the crucial role of band selection as a foundational preprocessing step. This step is essential to enable more precise and computationally efficient remote sensing analysis~\cite{sun2019hyperspectral}.

These methodologies can be further classified based on the nature of the processed data—whether labeled or not—into three categories: supervised, semi-supervised, and unsupervised methods ~\cite{7152840,6977960}. Supervised and semi-supervised approaches rely on labeled data to perform the selection process. However, acquiring labeled remote sensing data is often prohibitively expensive~\cite{10034816}, rendering supervised and semi-supervised methods impractical for widespread application. Consequently, this research prioritizes an unsupervised approach to band selection, using deep learning techniques to navigate the complexities of hyperspectral data without the need for labeled datasets~\cite{9387453,8715383}.

Recent advances have demonstrated the potential of combining deep learning with clustering or ranking methods for band selection in hyperspectral imaging (HSI), resulting in significant improvements~\cite{8356741,ayna2023learning}. The success of these integrations can be attributed to the adaptability of deep learning algorithms to identify and learn the intricate, nonlinear spatial and spectral relationships within HSI data~\cite{feng2019learning}. This innovative blend represents a substantial advancement in the field and combines the power of deep learning to refine band selection processes.

In the realm of recent studies on band selection, some researchers have redefined the task as one of reconstruction~\cite{8517884,9082156}. This perspective advocates for the sparse reconstruction of Hyperspectral Imaging (HSI) bands using a carefully chosen subset of informative bands~\cite{10189843}. This approach diverges from traditional unsupervised band selection methods by focusing on identifying bands that offer the highest level of informativeness and dissimilarity~\cite{9955115}. This strategy reflects a shift towards viewing hyperspectral image processing as a feature extraction ~\cite{feng2017hyperspectral,9952114,8738051} and selection challenge. The emphasis on maximizing the informativeness and dissimilarity of the selected bands is consistent with previous findings~\cite{4656481}, which highlight the importance of these criteria in improving the accuracy and efficiency of hyperspectral image analysis~\cite{feng2017hyperspectral,9952114}.

To use the complementary information provided by the LiDAR data, numerous studies have been carried out~\cite{feng2021hyperspectral,huang2023attention}.  These include the introduction of Progressive Extinction Mechanisms (PEM) for the fusion of Hyperspectral Imaging (HSI) and LiDAR data~\cite{rasti2017hyperspectral,tang2023active},aiming to extract both spatial and elevation information. This approach facilitates the creation of functional classification maps and improves piecewise smoothness while preserving spatial structure. Furthermore, Falahatnejad and Karami~\cite{falahatnejad2022deep} have successfully integrated the spectral and spatial content of HSI with the height detection information from LiDAR through a 2D Convolutional Neural Network (2DCNN) architecture, yielding promising results in classification improvement. Subsequent research has expanded on this foundation, employing attention-guided~\cite{huang2023attention} and coupled adversarial learning for fusion~\cite{lu2023coupled}, thus enhancing the classification accuracy of hyperspectral images and LiDAR data. However, there is a noticeable gap in the literature regarding the combination of hyperspectral image band selection with LiDAR, with the aim of classification enhancement.

Motivated by unsupervised band selection methods and inspired by  a Lidar guided fusion model~\cite{yang2024lidar} for supervised band selection, we intended to develop a unsupervised band selection deep learning model which can take the advantage of LiDAR attention scroes to improve hyperspectral image classification performance. THerefor a new unsupervised learning framework is proposed, named the Dual-Attention-Based Fused Mask Autoencoder. This framework is tailored to improve the selection of the hyperspectral band by integrating HSI and LiDAR data. It utilizes a novel dual-attention mechanism to analyze and pinpoint the most informative bands, exploring both the spectral and spatial dimensions of the data. This mechanism is bifurcated into two components: one focusing on the spectral correlations among the HSI bands and the other on the spatial correlations indicated by the LiDAR data. By synthesizing attention masks derived from both HSI and LiDAR inputs, one guarantees a holistic band selection. The framework culminates in a clustering phase, steered by an innovative distance metric, to isolate the most indicative subset of bands effectively. This method promises to significantly enhance band selection, thus optimising spectral and spatial analysis capabilities essential for remote sensing applications.

In this paper, we outline our contributions as follows:
\begin{itemize}
  \item Innovative Framework Employing LiDAR Data for Enhanced Band Selection: We introduce a novel architectural framework that leverages LiDAR data to enhance the feature representation of Hyperspectral Imaging (HSI) bands. This was achieved by creating a fusion mask that combines spatial information from LiDAR with spectral data from HSI, providing a richer and more detailed representation for further analysis.

  \item Fusion of Attention Mechanism with Autoencoder for Efficient Representation: Our architecture integrates an attention mechanism with an Autoencoder, ingeniously designed to capture a compact yet highly informative representation of the merged HSI and LiDAR data. A key component of this architecture is a construction loss that includes a term to induce sparsity, ensuring that the model focuses on the most critical features of both types of data.

  \item Customised Distance Metric for Superior Band Selection: We propose a unique distance metric specifically tailored to measure dissimilarity and pinpoint bands with the highest information content. This metric cleverly incorporates attention scores into its calculations, reducing the effective distance for bands with higher attention scores to prioritize their selection during clustering.

  \item Targeted Band Selection via Hierarchical Clustering: Utilizing hierarchical clustering and our bespoke distance metric, our method precisely selects bands that are not only individually significant due to high attention scores, but also exhibit minimal similarity to each other. This deliberate selection strategy ensures that the chosen bands offer a comprehensive and varied feature set, crucial to the intended analytical objectives.
\end{itemize}

The remainder of this paper is organised as follows. Section~\ref{sec:relatedwork} reviews related work in the literature. Our method is described in Section~\ref{sec:methodology}, followed by the experiments and results presented in Section~\ref{sec:experiments}.  Finally, Section~\ref{sec:conclusions} concludes this article.
 
\section{Related Work}~\label{sec:relatedwork}
This section provides a comprehensive overview of the unsupervised band selection methods and their integration with LiDAR data in hyperspectral image processing models. These methods, crucial for efficiently reducing data dimensionality while preserving essential information, are categorized into ranking-based, clustering-based, attention-based, and fusion techniques. This overview not only outlines the methods but also explores their integration with Light Detection and Ranging (LiDAR) data to enhance HSI processing models.

\subsection{Unsupervised Band selection}
Unsupervised band selection is essential for managing the complexity of hyperspectral data, categorically split into ranking based and clustering based approaches. Ranking-based strategies assign a significance score to each band, selecting those with the highest scores through various methods, including information~\cite{8738051} , similarity~\cite{4656481}, sparse representation~\cite{6116223}, and distance-based criteria. These methods aim to prioritize bands based on their informational content, similarity to other bands, sparse coding techniques, or their distance/dissimilarity from other bands~\cite{morales2021hyperspectral} , thereby reducing redundancy and enhancing discriminative power.

Clustering-based methods, in contrast, group bands into clusters, selecting the most representative band from each to best represent data subspaces~\cite{8949559}, high-density regions~\cite{8026133}, or contribute to a sparse non-negative representation of the data. This approach streamlines data by ensuring a broad yet concise representation through selected bands~\cite{li2011clustering}.

The ranking-based methods further delves into advanced techniques employing deep learning to refine band selection. These include methods like Similarity-Based Ranking with Structural Similarity (SSIM)~\cite{xu2021similarity}, Contrastive Learning for Band Selection (ContrastBS)~\cite{li2023unsupervised}, deep reinforcement learning~\cite{9387453}, and Band Grouping with Multigraph Adaptive Constraint~\cite{you2022hyperspectral}. Each method applies innovative criteria and learning models to enhance selection efficiency, reduce redundancy, and improve classification accuracy by focusing on structural similarities, mutual information maximization, sequential selection for reward maximization, and spectral correlation, respectively.
 Despite their effectiveness in improving analysis efficiency and accuracy, these methods also present challenges, notably the computational demands of detailed image generation required by some approaches. This comprehensive overview sets the stage for further exploration of these techniques, especially in how they integrate with other technologies like LiDAR to push the boundaries of hyperspectral image processing..

The examination of clustering based methods within hyperspectral image (HSI) analysis reveals their critical role in mitigating the challenges of data dimensionality and redundancy. By dividing the full spectrum of bands into distinct clusters and selecting a representative band from each, these methods streamline the process of hyperspectral imaging, ensuring that the essential discriminative information is retained.
Recent innovations have introduced a cluster-based framework leveraging optimal transport theory ~\cite{wang2022hyperspectral}, as described in select studies. This framework excels in producing optimal clustering outcomes within given constraints and incorporates a cluster classification approach along with an automated system for determining the optimal band count~\cite{Caron2020UnsupervisedLO}. Such advancements signify a move towards preserving critical information more efficiently.

In essence, the evolution of clustering-based methods in hyperspectral imaging is marked by increasingly sophisticated algorithms and parameter optimization strategies. These developments not only augment the precision and efficiency of band selection but also bolster the ability to extract and conserve crucial data, thus markedly advancing hyperspectral image analysis.

Partition-based strategies in hyperspectral band selection are emerging as effective solutions to the challenges of high dimensionality and data redundancy. These methods aim to divide the hyperspectral image cube into sub-cubes, selecting bands that encapsulate the most informative and distinct features. Strategies like the Adaptive Subspace Partition Strategy (ASPS)~\cite{8854164} excel by maximizing the distance ratio between pixel classes, selecting minimally noisy bands, and thus reducing redundancy while conserving information.

Moreover, methods like RLFFC~\cite{wang2022hyperspectral} introduce fusion-based clustering sensitive to regional features, employing superpixel segmentation to maintain spatial data integrity in HSIs. This approach enhances band separability and redundancy capture, using k-means clustering for band selection and showing notable improvements on public datasets. The Curve Fitting Subspace Partitioning (CFSP)~\cite{10382646} method segments hyperspectral bands into subcubes based on maximum curvature points, grouping similar and adjacent bands to minimize correlation among selected bands. It achieves high classification accuracy with minimal bands across several datasets, illustrating the robustness and efficiency of partition-based methods in hyperspectral band selection.

Attention-based techniques in hyperspectral imaging utilize attention mechanisms to identify the most informative bands, offering a novel approach to band selection. Methods like the Attention-based Autoencoder (AAE) \cite{dou2020band}focus on sparsity and global band interdependencies, employing attention modules for band selection and clustering, albeit with challenges related to computational demands and method clarity.

Another approach involves using clustering attention masks for band selection, demonstrating the capacity of attention-based methods to handle complex and non-linear band relationships, reducing redundancy, and enhancing post-training efficiency despite longer GPU training times~\cite{liu2022band}.

In summary, attention-based methods present a promising avenue for hyperspectral band selection, capable of addressing intricate band interdependencies. Nonetheless, they confront obstacles in computational efficiency, parameter tuning, and the clarity of the attention mechanisms employed.

\subsection{ HSI and LiDAR Fusion Models}
Moving forward with the fusion model, we would like to investigate these unspervised fusion models and supervised fusion model based on their architecture and label dat used. 

Regarding the supervised HSI and LiDAR fusion model, in particular, 
architectures based on convolutional neural networks (CNN) have been
instrumental in the fusion of HSI and LiDAR data, offering robust
spatial feature extraction capabilities~\cite{8985546}. Falahatnejad
et al.~\cite{falahatnejad2022deep} used hybrid 3D/2D CNNs and attention
modules to extract spatial spectral characteristics of HSI and
representative elevation characteristics of
LiDAR.EndNet~\cite{hong2020deep}, an encoder-decoder fusion network,
addresses the limitations of single-modal remote sensing data and uses
a reconstruction strategy to activate neurons in both modalities. 

Regarding unsupervised learning, the collaborative contrastive learning method (CCL)~\cite{jia2023collaborative} uses a pre-training and fine-tuning strategy to extract features from HSI and LiDAR data separately and produce a coordinated representation and matching of features between the two-modal data without label samples.
 The FusAtNet model~\cite{mohla2020fusatnet} fuses self-attention and cross-attention modules, highlighting the harmonious integration of the HSI and LiDAR data.  The extracted features were then integrated and classified using a softmax classifier.
In~\cite{xu2023unsupervised}, authors introduce semantic understanding to dynamically fuse data from two different sources, extract features of HSI and LiDAR through different branches of the capsule network, and improve self-supervised loss and random rigid rotation in the canonical capsule to a high-dimensional situation. 

Deep learning and attention-based fusion techniques take advantage of the strengths of both types of data, providing comprehensive insight that is crucial for various applications. However, there is a noticeable gap, as current approaches tend to use full HSI bands and LiDAR data without performing any band selection.  

In summary, attention-based methods are a promising direction for band selection in hyperspectral images, as they can capture complex and nonlinear relationships between bands and reduce redundancy. However, they also face some challenges, such as computational efficiency, parameter tuning, and interpretability of the attention mechanism.

After investigation, when multiresource data is available, fewer researchers considered using more external features for the selection of hyperspectral image bands. Therefore, we plan to propose multi-data unsupervised band selection to improve the hyperspectral image data band selection and increase the downstream task.  A cluster integration-dual-attention-band selection method is proposed for hyperspectral image band selection with the aid of LiDAR data.

\section{Methodology}\label{sec:methodology}
This section describes our proposed methodology for the selection of unsupervised hyperspectral bands, integrating a dual attention mechanism with an Autoencoder framework. Our methodology is structured around two core components: the generation of attention masks from both HSI and LiDAR data, and the subsequent application of these fused masks to the combined HSI and LiDAR inputs in an Autoencoder for band selection. This approach conceptualises band selection as a reconstruction challenge, aiming to retain a representative subset of bands that closely mirror the original hyperspectral data, once fused with LiDAR information. 

\subsection{Dual Fused Mask Attention-Autoencoder Framework for Band Selection Overview}
\begin{figure*}[t]
\centering
\includegraphics[width=17cm,height=7cm]{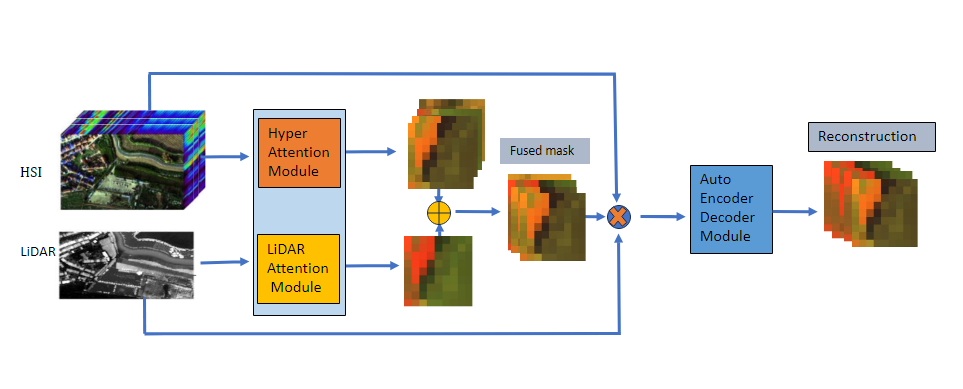}
\caption{ Flowchart of the HSI band Selection }
\label{fig_1}
\end{figure*}

Dual Fused Mask Attention-Autoencoder Framework for Band Selection.
The architecture of our framework, shown
in Fig. \ref{fig_1}, describes the original hyperspectral data post-fusion with LiDAR, as depicted in the corresponding flow chart. At the heart of this framework lies a dual Attention Autoencoder (AAE) network designed to process the fused HSI and LiDAR data inputs effectively. Using an unsupervised convolutional neural network (CNN) paradigm, the Autoencoder strives to replicate its input within its output closely, facilitating the selection of the most informative and minimally redundant bands. This selection is refined through a custom distance function that guides the hierarchical clustering method to isolate the top ${k}$ bands of interest.

Our framework's uniqueness lies in its dual-attention module, embedded at the foundational level to discern the most salient bands within the dataset. This module bifurcates into the HSI attention mechanism, focusing on extracting nonlinear correlational attention among hyperspectral bands, and the LiDAR attention mechanism, which unravels spatial correlations, enriching spectral analysis with spatial context.

Following attention analysis, each mechanism generates a sparse mask, the HSI mask, and the LiDAR mask, which are then concatenated. This combined mask undergoes a dot product operation with the input of the attention module, the result of which is fed into the Autoencoder. For the final band selection phase, the fused attention masks are subjected to a clustering algorithm, steered by a novel distance metric designed to act as the selection criterion. This metric is critical to select a band subset that effectively encapsulates both the spectral and spatial attributes.

To facilitate understanding, we introduce the specific notation used throughout this discussion.
Let ${( X_{hsi} \in \mathbb{R}^{H \times W \times B})}$ be an HSI with $(B)$ bands and ${H \times W}$ pixels, $ X_{lid} \in \mathbb{R}^{H \times W}$. In the data pre-processing step, a square window of size $p \times p$ slides across the original HSI and LiDAR to obtain a sufficient number of spatial spectral HSI patches $ X_{p} \in \mathbb{R}^{p*p, B}$ as input to the dual attention network. Suppose that the sample data obtained by flattening each sample from the original HSI band into a vector is denoted as $ {X_{hsi2D} = [x_{hsi1}, x_{hsi2}, \ldots, x_B] \in \mathbb{R}^{P \times B}}$, where $B$  represents the number of patches. The objective of our method is to select the desired subset of bands consisting of \( k \) bands, and a candidate band subset can be expressed as \( X_s = [x_{s(1)}, x_{s(2)}, \ldots, x_{s(k)}] \in \mathbb{R}^{P \times k} \).

Through this methodology, our aim is to introduce a novel and efficient approach to hyperspectral band selection, leveraging the complementary strengths of HSI and LiDAR data to enhance the accuracy and efficiency of remote sensing data analysis.

\subsection{Proposed Methods Implementation}

\begin{figure*}[!t]
\centering
\includegraphics[width=17cm,height=10cm]{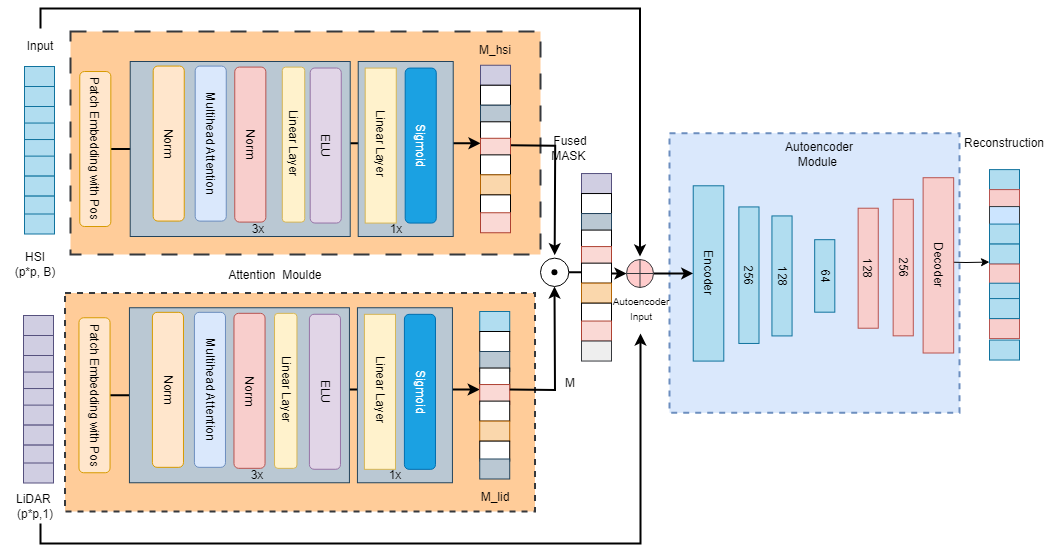}
\caption{Detailed Architecture of HSI and LiDAR MASK fusion based on Attention Module for HSI band Selection and Autoencoder Reconstruction.}
\label{fig_2}
\end{figure*}

Fig. \ref{fig_2} shows the detailed dual attention and Autoencoder architecture. 

Given the architecture that fuses the HSI and light detection and
positioning (LiDAR) data, the objective function of training is to find
a sparse fused mask${M_{fused}}$ from HSI and LiDAR, respectively, 
${M_{hsi} \in \mathbb{R}^{P \times B}}$ and ${M_{lid} \in \mathbb{R}^{P 
\times 1}}$. where $P$ is the size of the patch and $B$ is the number of spectral bands. The model aims to learn a mapping function $f$,

The transformer has revolutionised data processing by using self-attention mechanisms to model long-range dependency and capture features in data~\cite{vaswani2017attention}.  Patch embedding is created based on divided patches with positional encoding, which can ensure the capture of the important feature of the context and retain the band position during data processing. Taking advantage of these advances,
dual self-attention attention modules are designed to generate HSI and LiDAR masks, respectively. Our dual-attention module for HSI and LiDAR consists of four fully connected layers. The first three layers are followed by ELU activation functions, and the last layer is followed by a Sigmoid function; detailed structure is illustrated in~\ref{fig_2}. Once HSI and LiDAR attention masks are obtained, the LiDAR mask is multiplied by the HSI mask to enhance the HSI mask attention, and then the fused HSI mask is multiplied by the raw HSI in an element-wise manner to be the input of the Autoencoder. Equation~\ref{eq1} is the fused mask.
\begin{equation}
 X_{hsi} = f(X_{hsi} \odot M_{hsi}) 
\label{eq1}
\end{equation}

where ${\odot}$ is the elemental multiplication.

The mapping function $f$ uses the CNN Autoencoder.  The Autoencoder takes the masked HSI $X_{hsi}$ as input and outputs the reconstructed HSI. The Autoencoder consists of four layers with filter (3, 3), pooling (2,2), and ElU activation function.  According to the description above of the dual attention module and the Autoencoder module, the attention module captures the interdependencies between bands and can obtain the attention mask of one sample and batch samples. 
Taking into account the fusion of HSI and LiDAR, the objective function~\ref{eq3} can be updated as follows through model training.

\begin{equation}
\begin{split}
J(\theta) = \frac{1}{2} \left\| f\left((X_{\text{HSI}} \odot M_{\text{fused}}) \right. \right. ) - X_{\text{HSI}} |_2^2 + \lambda \left\| M_{\text{fused}} \right\|_{2,1} 
\label{eq3}
\end{split}
\end{equation}

where \begin{itemize}
    \item \( X_{\text{HSI}} \) and \( X_{\text{LiDAR}} \) are the input data from the HSI and LiDAR sensors, respectively.
    \item \( M_{\text{HSI}} \) and \( M_{\text{LiDAR}} \) are the sparse masks for the HSI and LiDAR data.
    \item \( \odot \) denotes element-wise multiplication.
    \item \( f \) is the mapping function learned by the network.
    \item The term \( \left\| \cdot \right\|_2^2 \) denotes the squared Euclidean norm, indicating the reconstruction loss.
    \item The term \( \left\| \cdot \right\|_{2,1} \) denotes the \( L_{2,1} \) norm which promotes sparsity in masks.
    \item \( \lambda \) is a regularization parameter that balances the accuracy and sparsity of the mask reconstruction.
   \item $M_{fused}$ is $M_{hsi}$ and $M_{lidar}$ by adding the operation, which is defined as the square root of the sum of the squares of the L1 norms of the rows of the matrix M, which can be used to promote the sparcity of the vector of rows M during the training of the objective function.
\begin{equation}
\left\|M_{fused}\right\|_{2,1} = \sqrt{\sum_{i=1}^{n} \| \mathbf{M}_{i, :} \|_1^2}
\label{eq2}
\end{equation}
\end{itemize}

\subsection{Custom Score Function for Band Selection}
In this study, we introduce a new custom distance function as a key criterion for band selection in hyperspectral imaging. This function is meticulously designed to integrate three core components: the normalised attention scores for each band (referred to as $A_{norm}$ in Equation~\ref{eq6}, the dissimilarity measure derived from the Pearson correlation matrix (defined in $D_{dissimilarity}$, Equation~\ref{eq7}). These components collectively facilitate a multifaceted assessment of band importance, inter-band similarity, and their joint contribution to the overall information content.

Initially, the function computes a distance matrix that encapsulates two fundamental types of information: the similarity between bands, as indicated by the Pearson correlation matrix, and the significance of each band, as reflected by its normalised attention score. The distance associated with each band is inversely proportional to its attention score, highlighting the importance of high-attention bands. On the contrary, the distance is directly proportional to the dissimilarity, which is computed as one minus the correlation coefficient, highlighting the uniqueness of each band.

To adeptly balance the influence of band importance (attention) and distinctiveness (dissimilarity), the distance measure integrates these aspects using adjustable parameters $\alpha$ and $\beta$  .Here, $\alpha$ modulates the weight given to the attention-based component, while $\beta$  adjusts the emphasis on dissimilarity. This balanced approach is crucial to tailoring band selection to both data-driven requirements and task-oriented relevance criteria. The bands thus selected are expected to be highly informative for subsequent analyses in hyperspectral imaging contexts, embodying an optimal blend of data representation and attention-derived significance.

The equations governing the custom distance function are delineated below.

\subsubsection{Aggregate Attention Values}
Compute the mean attention value in all samples and spatial locations for each band, assuming the batch size is (N) and the sample patch size is (p). The computing equation is as follows:
\begin{equation}
A_i = \frac{1}{N \times p \times p} \sum_{n=1}^{N} \sum_{x=1}^{p} \sum_{y=1}^{p} M_{n,x,y,i}
\label{eq4}
\end{equation}
This results in a vector \(A\) of length \(B\), where each \(A_i\) is the aggregated attention score for the \(i\)-th band of \(B\).

\subsubsection{Normalize Attention Values}
Normalize the aggregated attention scores to ensure that they are on a comparable scale:
\begin{equation}
A_{\text{norm}} = \frac{A_i - \min(A)}{\max(A) - \min(A)}
\label{eq5}
\end{equation}

\subsubsection{Attention Distance Matrix}
Based on normalised attention values, define the attention distance matrix \(D_{\text{attention}}\) where each element \(D_{\text{attention},i,j}\) represents the combined attention distance for bands $i$  and $j$. This attention matrix is computed using the outer product of the attention vector as below. 

\begin{equation}
D_{\text{attention},i,j} = A_{\text{norm},i} \times A_{\text{norm},j}
\label{eq6}
\end{equation}
The attention  matrix captures the relative attention-based separation between each pair of bands, with higher values indicating high attention.

\subsubsection{Dissimilarity Distance Matrix}
Let \(D_{\text{dissimilarity}}\) be the dissimilarity matrix where each element \(D_{\text{dissimilarity},i,j}\) is defined based on the Pearson correlation between bands \(i\) and \(j\):
\begin{equation}
D_{\text{dissimilarity},i,j} = 1 - \text{Corr}_{i,j}
\label{eq7}
\end{equation}
where \(\text{Corr}_{i,j}\) is the Pearson correlation coefficient between bands \(i\) and \(j\). Bands with high correlation may provide redundant information, leading to lower dissimilarity values.

\subsubsection{Combined Distance Matrix for Clustering}
To facilitate the clustering process, we define \(D_{\text{combined}}\) as the matrix that represents the combined distance. This matrix integrates both spectral and spatial distances, enabling a more nuanced band selection process. 
 The calculation of the combined distance is as follows:

\begin{equation}
D_{\text{combined},i,j} = \alpha \times D_{\text{attention},i} + \beta \times D_{\text{dissimilarity},i,j}
\label{eq9}
\end{equation}

where $\alpha$ and $\beta$ are adjustable parameters that control the influence of the distance of attention, the distance of dissimilarity and the weight matrix, respectively. These parameters should sum up to 1, allowing flexibility in the band selection process.

Subsequently, the band selection process involves performing hierarchical clustering using \( D_{\text{combined}} \) to group bands into clusters. A subsequent algorithm, k-means, is then applied to select the top \( k \) bands of these clusters. For each cluster, the band with the highest attention score norm \( A_{\text{norm}} \) is selected. This ensures that the selected bands are representative of the variability of the data set and are deemed the most important by the attention mechanism.

\section{Experiments}\label{sec:experiments}
This section covers the data set used in our experiments, the experimental setup, detailed results of the tests and an in-depth analysis of the ablation study. This thorough examination aims to provide a holistic understanding of our method's performance and its practical implications in the field of hyperspectral imaging.

\subsection{Data sets Collection and Pre-processing}
We have carried out extensive experiments to assess the performance of the proposed cross-attention band selection method. The experiments were run on three paired HSI and LiDAR datasets, Houston 2013, Trento, and MUUFL.

 Houston 2013 dataset contains hyperspectral and LiDAR data for the IEEE 2013 GRSS Data Fusion~\cite{houston_dataset_2013}. It includes a hyperspectral image (HSI) with 144 spectral bands (380-1050nm) and a LiDAR-derived digital surface model (DSM), both at 2.5 m resolution. The HSI is calibrated to the radiance of the sensor, while the DSM indicates elevation in meters above sea level. The data set contains 15 types of land cover, making it ideal for evaluating band selection and classification methods, despite urban complexity and challenges related to noise in the HSI. In our experiments, we follow the number of standard training data to perform the final classification test and comparison.

The Trento data set~\cite{university_of_trento_theses_2022}  was acquired in Trento, Italy, in 2015. The hyperspectral image comprises 48 spectral bands (400-950 nm) with a one-meter spatial resolution, whereas the LiDAR data present the DSM. The data set covers approximately 100 hectares and includes six land cover classes. 

The MUUFL Gulfport scene\cite{du_zare_2017}contains hyperspectral and LiDAR data collected over Gulfport, Mississippi, in 2010 and 2011. The data set includes four subimages with different spatial resolutions and elevations and ground truth information.  There are 11 classes in this data set. 
For the experiments consistently, we used the standard test samples in the final classification and co-comparative experiments. 

Given that our band selection method operates in an unsupervised manner, it does not incorporate ground-truth labels during the extraction of training samples. To validate classification performance, we employ the following approach in comparative analysis with alternative methods to assess its efficacy.

\subsection{Competitors and Experiment Configurations }

Competitors are categorised into two groups. The methods selected for group one are band selection algorithms. The other group chosen is for fusion models. 

\subsubsection{Competitors of Band Selection Methods}
In our study, we employ an optimally trained model to perform a comprehensive comparative analysis of various state-of-the-art unsupervised hyperspectral band selection methods. Each comparison method is implemented and evaluated according to the specific descriptions and parameters outlined in their respective original publications. This approach ensures a fair and consistent comparison across all methods.

SR-SSIM~\cite{xu2021similarity}  uses structural similarity indices to analyse band relationships. ASPS~\cite{8854164} employs an adaptive subspace partition strategy to improve the discriminability of the band.
TOF~\cite{8356741}focuses on optimal clustering with a unique rank-on-cluster strategy for band selection. RRI~\cite{liu2022band} estimates the representativeness of the band through a convolutional Autoencoder, balancing redundancy and information content. RLFFC~\cite{wang2022hyperspectral} uses superpixel segmentation and latent feature fusion for hyperspectral images, improves spatial and spectral data, and uses clustering to reduce band redundancy.
These unsupervised band selection methods are included in our band selection comparative study. These selected bands from different methods are integrated with LiDAR to have a comprehensive evaluation of classification performance. 

\subsubsection{Competitors of Fusion Models }
In recent years, deep learning has been applied to build fusion models in the field of remote sensing. In this selection, we explore the influence of the application of the BS method integrated with the fusion model. 
Both the SVM and 2DCNN deep supervised learning classifier is integrated with our unsupervised BS method to have a comparison with other fusion models. 

The fusion models selected for this comparative study include supervised deep learning methods, the CNN-HSI model proposed by Yu et al. (2017)~\cite{yu2017convolutional} and further developed by Mohla et al. (2020)~\cite{mohla2020fusatnet} , the CoupledCNN model of Hang et al. (2020)~\cite{hang2020classification},  and a general multimodal deep learning framework (MDL-RS) including early fusion, middle fusion, and late fusion models explored by Hong et al. (2020)~\cite{hong2020more},  additional comparisons made with EndNet (2020)~\cite{hong2020deep}.  Regarding unsupervised learning, include the collaborative convolutional learning model (CCL) presented by Jia et al. (2023)~\cite{jia2023collaborative}, FuseAtNet as detailed by Mohla et al. (2020)~\cite{mohla2020fusatnet}.

These state-of-the-art fusion models typically operate on full-band HSI hyperspectral image data combined with LiDAR. All experimental comparisons adhere to the same test settings as those defined in their rearch. Instead, our method uses only 10 HSI bands from the unsupervised method and fuses them with the LiDAR data. 

\subsubsection{Experimental Configuration and Parameters Setting}
To evaluate the performance of our proposed band selection method (BSFMAA) and other competitors, we chose two classifiers, including SVM and KNN. In the experiment, the parameter {K} of KNN classifier is set to 5, and rbf is the kernel of the Radial Basis Function (RBF) used in SVM. All tests are performed on Google Colab TPU. 

In detail, we systematically assess the classification efficacy of our proposed band selection method from multiple perspectives. Our proposed model consists of dual attentions, fused masks from dual attention architectures, and four-layer convolutional networks for the encoder and decoder, respectively. Hyperparameters for the training model include Epoch 50, SGD Optimizer, batch size 32, the learning rate defined as 0.0001, and Optimiser is SGD. 

For band selection evaluation, we use consistent classification tools, namely the Support Vector Machine (SVM) and the K-Nearest Neighbour (KNN). These classifiers are applied to the same ratio of training and test data for the assessment of band selection methods.

For further fusion model evaluation, we use 10 selected bands that integrate SVM and 2DCNN as our fusion methods to compare these state-of-the-art fusion models.

\subsubsection{Accuracy Assessment}
 Key metrics for this evaluation include classification accuracy and the number of bands required. Higher accuracy and fewer bands are indicative of improved discrimination power and a more efficient reduction in the dimensionality of hyperspectral images. The band number ranges from 1 to 50 and the interval is 5. Three common indexes, namely overall precision (OA), average precision (AA), and Kappa coefficient (Kappa), are used to have a qualitative comparison of classification performance based on different band selection methods.  All test and training samples are the same as our proposed method and companions for suitability.

\subsection{ Band Selection Experimental  Results and  Analysis}
We conducted extensive experiments using three different data sets to identify the model that performs optimally with specific configurations. Our approach integrates a dual-attention mechanism with a fused mask. We tested various patch sizes (3, 5, 7, 9, 11, and 13) and hyperparameters to determine their impact on performance. The combination of a patch size of 7, a batch size of 32, the SGD optimizer, and a learning rate of 0.0001 yielded the best results. To ensure fairness in our evaluation, we used the SVM and KNN classifiers and 5 layers of 1DCNN to assess classification performance. Our baseline comparison involves using standard SVM and KNN algorithms with full-band HSI and LiDAR data as input.

\subsubsection{UH2013 Experiments}
\begin{figure}[htp!]
\centering
\includegraphics[width=8cm, height=5cm]{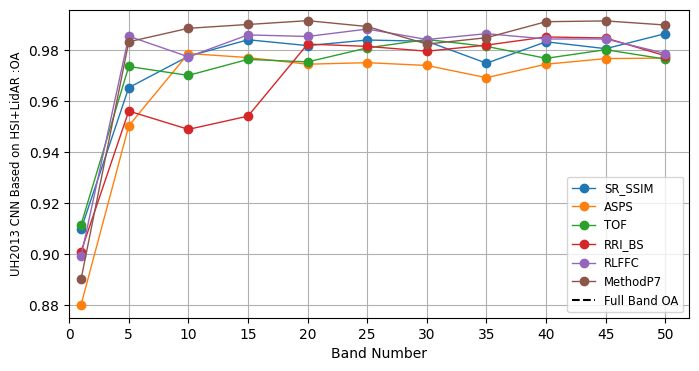}
\label{uh_cnn_oa}
\caption{CNN OA  Comparison-Based on UH2013 Data set using Different Band Selection Methods.}
\label{fig_uh_cnn}
\end{figure}

Table~\ref{uh2013_bs_results} shows the 10 bands OA, AA and Kappa of all methods in the Houston 2013 data set. The outcome based on the data set revealed that our method improved overall precision (OA), average precision (AA) and Kappa metric by approximately 2\%, 1. 5\% and 1. 5\%, respectively, in the SVM evaluations. In the KNN assessments, we observed an improvement 2\% in both OA and AA. The results underscore the effectiveness of our method, highlighting its superior performance against competing approaches, as detailed in the table below, which compares the results across various methods using the UH2013 dataset. Regarding the CNN assessment, it still shows that our selected bands are in the best place~\ref{fig_uh_cnn}.

\begin{table*}[tp]
\centering
\caption{UH 2013 Three Classification Test Results Based on Fewer Band Number 10.\\ \textcolor{blue}{Blue Represents the Best method}, \textcolor{red}{Red Represents the second place model}.}
\label{tab:uh2013_results}
\begin{tabular}{|l|l|l|l|l|l|l|l|l|}
\hline
\hline
Method & Metrics &BaseLine(H+L) & SR\_SSIM~\cite{xu2021similarity} & ASPS~\cite{8854164} & TOF~\cite{8356741} & RRI\_BS~\cite{liu2022band} & RLFFC~\cite{wang2022hyperspectral}& OurMethod \\ 
\hline
\hline

\multirow{3}{*}{SVM} 
& OA & 0.8960 &0.8598 &0.8614 & 0.8282& 0.7635  &  \textcolor{red}{0.9007}& \textcolor{blue}{0.9228} \\ 
& AA & 0.9073 & 0.8807 & 0.8776 & 0.8515 & 0.7992 & \textcolor{red}{0.9162} & \textcolor{blue}{0.9308} \\ 
& Kappa & 0.8872 & 0.8479 & 0.8496& 0.8138& 0.7443 & \textcolor{red}{0.8923} & \textcolor{blue}{0.9162}\\ 
\hline

\multirow{3}{*}{KNN} 
& OA & \textcolor{red}{0.8852}& 0.8540 & 0.8475 & 0.8083& 0.7920 & 0.8737 & \textcolor{blue}{0.8986} \\
& AA & \textcolor{red}{0.8633} &0.8355 & 0.8327 & 0.7991 & 0.7848& 0.8495& \textcolor{blue}{0.8755} \\
 & Kappa & \textcolor{red}{0.8753} & 0.8414& 0.8345 & 0.7921 & 0.7746 & 0.8629 & \textcolor{blue}{0.8898} \\ 
 \hline
 
\multirow{3}{*}{CNN} 
& OA &\textcolor{red}{0.9790} &0.9341&	{0.9341}&	0.9447&	0.9490&	{0.9775}&	\textcolor{blue}{0.9887} \\
& AA &0.9806 &0.9418&	0.9418&	0.9489&	0.9587&	\textcolor{red}{0.9826}&	\textcolor{blue}{0.9912} \\
 & Kappa &\textcolor{red}{0.9772} &0.9284 &	{0.9284}&	0.9399&	0.9447 &0.9755&	\textcolor{blue}{0.9878}\\ 
 \hline
 \hline
\end{tabular}
\label{uh2013_bs_results}
\end{table*}

\begin{figure}[p]
\centering
\includegraphics[width=8cm, height=5cm]{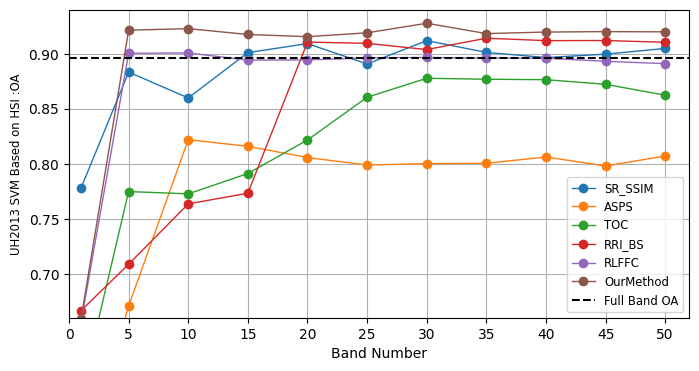}
\label{uh_svm_oa}
\caption{SVM OA  Comparison-Based on UH2013 Data set using Different Band Selection Methods.}
\end{figure}

\begin{figure}[htp!]
\centering
\includegraphics[width=8cm, height=5cm]{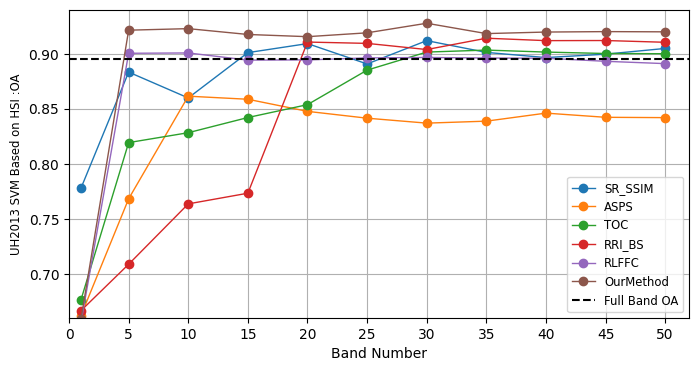}
\label{uh_knn_oa}
\caption{KNN OA  Comparison-Based on UH2013 Data set using Different Band Selection Methods.}
\end{figure}

Figs. ~\ref{uh_svm_oa} and~\ref{uh_knn_oa} depict classification performance across bands 1 to 50 (in intervals of 5) further illustrate our method's consistency and exceptional performance when evaluated using both SVM and KNN classifiers.

\subsubsection{Trento Experiments }
Similarly to the UH2013 results, our experiments with the Trento data set show the efficacy of our method in Table~\ref{trento_bs_results}.  It outperforms others by 0. 4\% in OA, 1. 4\% in AA, and 0. 8\% in Kappa when using SVM. KNN results were even more pronounced, with improvements of 2. 5\% in OA and Kappa and 1\% in AA over the next best method. This demonstrates our approach's robustness and reliability across different datasets and classification techniques. Although the CNN result has a slight increase, some AA is not in the best place, in the band range 5-20, our method is the best~\ref{fig_mf_cnn}.

\begin{table*}[tp]
\centering
\caption{Trento Test Result Based on Fewer Band Number 10.\\ \textcolor{blue}{Blue Represents the Best method}, \textcolor{red}{Red Represents the second place model}}
\label{trento_bs_results}
\begin{tabular}{|l|l|l|l|l|l|l|l|l|}
\hline
\hline
Method & Metrics &BaseLine(H+L) & SR\_SSIM~\cite{xu2021similarity} & ASPS~\cite{8854164} & TOF~\cite{8356741} & RRI\_BS~\cite{liu2022band} & RLFFC~\cite{wang2022hyperspectral}& OurMethod \\ 
\hline
\hline
\multirow{3}{*}{SVM} 
& OA & 0.9820 & \textcolor{red}{0.9831} & 0.9813 & 0.9804& 0.9201& 0.9815 & \textcolor{blue}{0.9867}\\ 
& AA & 0.9712 & \textcolor{red}{0.9861} & 0.9726 & 0.9716 & 0.8953 & 0.9730 & \textcolor{blue}{0.9872} \\ 
& Kappa &0.9759&\textcolor{red}{0.9773} & 0.9748 &  0.9737 & 0.8907 & 0.9751 &\textcolor{blue}{0.9821} \\ 
\hline
\multirow{3}{*}{KNN} 
& OA & 0.9323 & 0.9431 &0.9519 & \textcolor{red}{0.9544} & 0.9227 & 0.9481 & \textcolor{blue}{0.9706}  \\ 
& AA & 0.9192 &0.9456 &0.9517 & 0.9448 & 0.9283& \textcolor{red}{0.9525} & \textcolor{blue}{0.9612} \\ 
& Kappa & 0.9103&0.9246 &\textcolor{red}{0.9360} & 0.9393& 0.8981 & 0.9312 & \textcolor{blue}{0.9608} \\ 
\hline
\multirow{3}{*}{CNN} 
& OA &0.9950 &\textcolor{red}{0.9958}&	0.9944&	0.9955&	0.9830&	0.9949&\textcolor{blue}{0.9962 }\\
& AA &0.9912&\textcolor{red}{0.9935}&	0.9918&	0.9901&	0.9805&	0.9920&\textcolor{blue}{0.9942} \\
 & Kappa &0.9933 &\textcolor{red}{0.9944}&	0.9926&	0.9940	&0.9772&	0.9933	&\textcolor{blue}{0.9949}\\ 
\hline
\hline
\end{tabular}
\end{table*}

\begin{figure}[htp!]
\centering
\includegraphics[width=8cm, height=5cm]{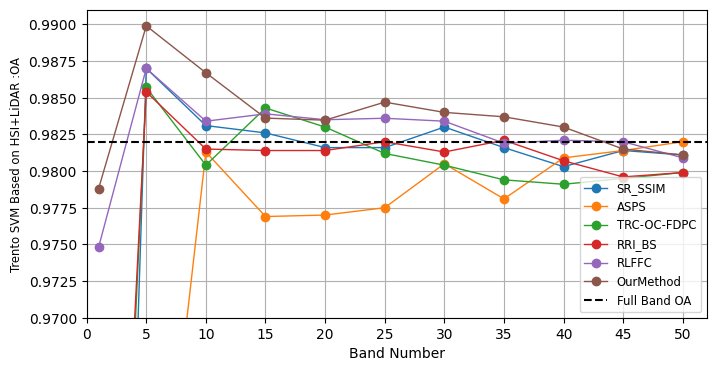}
\caption{SVM OA  Comparison-Based on Trento Data set using Different Band Selection Methods.}
\label{fig_tr_svm}
\end{figure}

\begin{figure}[htp!]
\centering
\includegraphics[width=8cm, height=5cm]{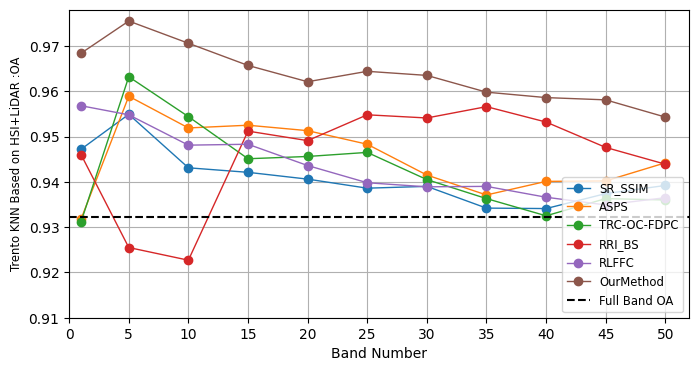}
\caption{KNN OA Comparison-Based on Trento Data set using Different Band Selection Methods.}
\label{fig_tr_knn}
\end{figure}

\begin{figure}[htp!]
\centering
\includegraphics[width=8cm, height=5cm]{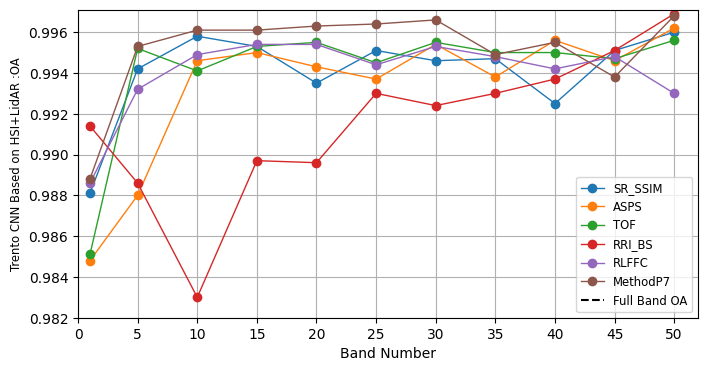}
\caption{KNN OA  Comparison-Based on Trento Data set using Different Band Selection Methods.}
\label{fig_tr_cnn}
\end{figure}

 Fig.~\ref{fig_tr_svm} and Fig.~\ref{fig_tr_knn} shows that the test results of our proposed method have superior performance and stability in the selected bands range from 1 to 50 interval 5.  For the outcome of the SVM, KNN and CNN OA test, our method is in the best place between 1 and 10, and between 15 and 20, our method is slightly lower $RRI_{BS}$, and then our method is resumed in the first place compared to other methods than in most cases.  
 Visual comparisons from the Trento dataset further affirm our method's superior classification accuracy and stability across selected bands, with our approach consistently leading in performance metrics.

\subsubsection{MUUFL Experiments}
In the exploration of the MUUFL dataset, particularly focusing on classification performance with a reduced set of five bands, our findings highlight a notable improvement in classification metrics. Our method shown in Table~\ref{muufl_bs_results}demonstrates an enhancement of 0. 5\% to 1. 5\% on the overall accuracy (OA), average accuracy (AA), and Kappa metrics for both the SVM and KNN classifiers over the second best performing methods.

\begin{table*}[tp]
\centering
\caption{MUUFL Test Result Based on Fewer Band Number 10. \\
\textcolor{blue}{Blue Represents the Best method}, \textcolor{red}{Red Represents the second place model}}
\label{muufl_bs_results}
\begin{tabular}{|l|l|l|l|l|l|l|l|l|}
\hline
\hline
Method & Metrics &BaseLine(H+L) & SR\_SSIM~\cite{xu2021similarity} & ASPS~\cite{8854164} & TOF~\cite{8356741} & RRI\_BS~\cite{liu2022band} & RLFFC~\cite{wang2022hyperspectral}& OurMethod \\ 
\hline
\hline
\multirow{3}{*}{SVM} 
& OA & \textcolor{red}{0.8477} & 0.8459& 0.8150& 0.8365 & 0.7594& 0.8451 & \textcolor{blue}{0.8591} \\ 
& AA & 0.8678 & 0.8596 & 0.8554 &0.8728& 0.8188 & \textcolor{red}{0.8680}& \textcolor{blue}{0.8766} \\ 
& Kappa &0.8008 &0.7814 & 0.7648& 0.7916& 0.6974& \textcolor{red}{0.8016} & \textcolor{blue}{0.8189} \\ 
\hline
\multirow{3}{*}{KNN} 
& OA & 0.7695& 0.7684 & 0.7993 & \textcolor{red}{0.8042}& 0.7371& 0.8035 & \textcolor{blue}{0.8097}\\ 
& AA & 0.8047 & 0.8186 & 0.8300 & \textcolor{red}{0.8337} & 0.7658 & 0.8328 & \textcolor{blue}{0.8353} \\ 
& Kappa &0.7096 &0.7300 & 0.7460 & 0.7521 & 0.6702 & \textcolor{red}{0.7514} & \textcolor{blue}{0.7586} \\ 
\hline
\multirow{3}{*}{CNN} 
& OA & \textcolor{red}{0.9017}& 0.8974&	0.8989&	0.8971&	0.8679&	0.8878&	\textcolor{blue}{0.9031}\\ 
& AA & \textcolor{blue}{0.9185}& \textcolor{red}{0.9064}&	0.8875&	0.8970&	0.8757&	0.9025&	0.9026 \\ 
& Kappa &\textcolor{red}{0.8725}&0.8666&0.8679&	0.8449&	0.8275&	0.8546&	\textcolor{blue}{0.8739 }\\ 

\hline
\hline
\end{tabular}
\end{table*}

\begin{figure}[htp!]
\centering
\includegraphics[width=8cm, height=5cm]{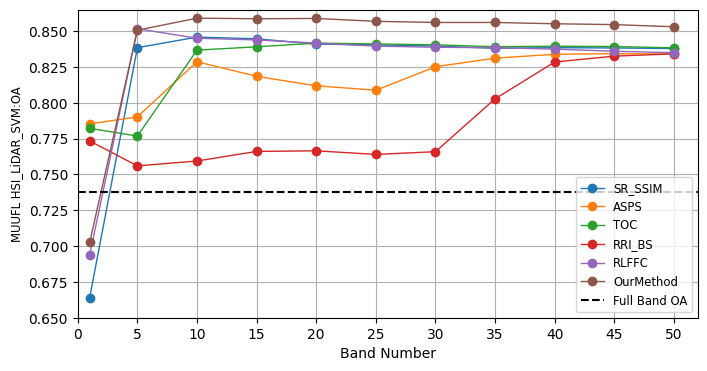}
\label{mf_svm_oa}
\caption{SVM OA Comparison-Based on MUUFL Data set using Different Band Selection Methods.}
\label{fig_mf_svm}
\end{figure}

\begin{figure}[htp!]
\centering
\includegraphics[width=8cm, height=5cm]{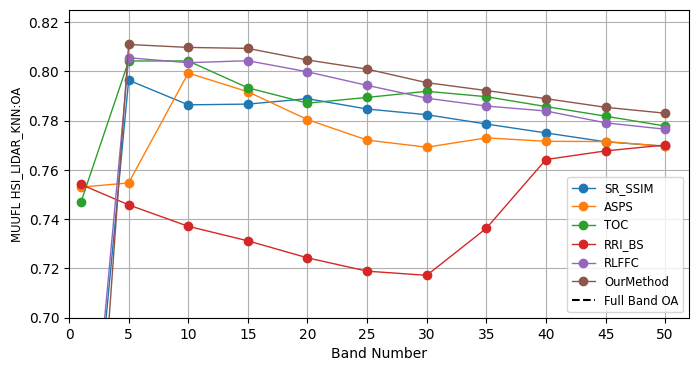}
\caption{KNN OA  Comparison-Based on MUUFL Data set using Different Band Selection Methods.}
\label{fig_mf_knn}
\end{figure}

\begin{figure}[htp!]
\centering
\includegraphics[width=8cm, height=5cm]{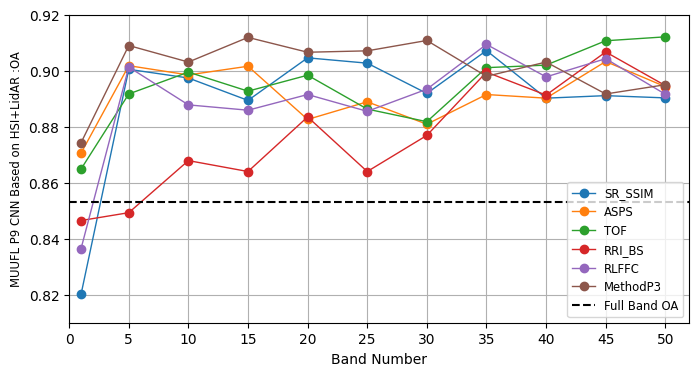}
\caption{CNN OA  Comparison-Based on MUUFL Data set using Different Band Selection Methods.}
\label{fig_mf_cnn}
\end{figure}

Fig.~\ref{fig_mf_svm}, Fig.~\ref{fig_mf_knn}, and Fig~\ref{fig_mf_cnn} show that our proposed method is clearly better than the other methods. For SVM, our approach can achieve the highest scores in OA outperforming the nearest competitor. Similarly, in the KNN and CNN evaluations, our method continued to lead, underscoring its robustness and effectiveness in handling hyperspectral image classification with a limited number of bands.

In summary, the experimental results of the three data sets demonstrate the efficacy of our proposed band selection method. Through rigorous testing and analysis, we have shown that our approach excels not only in enhancing classification metrics but also in ensuring stability and reliability across various classifier configurations and patch sizes.

\subsection{Fusion Models Experimental Results and Comparison }
This section delves into the evaluation of fusion models that incorporate our optimally selected bands with the SVM and a 5-layer CNN, respectively, forming a straightforward but effective fusion framework. We benchmark our models against supervised and unsupervised fusion approaches, with the fusion model introduced by Mercier et al.~\cite{mercier2003support}. serving as the primary baseline for comparison. The configurations of these selected fusion models adhere strictly to the definitions outlined in their respective studies.

\subsubsection{Fusion Comparison based on UH2013 Data Set}
A detailed comparison between the base model, our SVM fusion model, and our CNN model, integrating the selected 10 bands, is presented with a focus on 15 classes, overall accuracy (OA), average accuracy (AA), and Kappa coefficient. The SVM fusion model excels in 3 classes (C1, C4, C15), while the CNN fusion model leads in 7 classes (C3, C7, C8, C9, C10, C11, C12), showcasing the efficacy and precision of our fusion strategies across a broad spectrum of classes.

Table~\ref{uh_fusion_com}  illustrates a side-by-side performance metric comparison, highlighting where our methods (SVM and CNN fusion models) stand in relation to other advanced fusion methodologies. This clear delineation showcases our fusion models' competitiveness, particularly emphasising instances where our approaches outperform others, thereby reinforcing the effectiveness of our band selection and fusion technique. 

\begin{table*}[htp!]
\centering
\caption{CLASSIFICATION PERFORMANCE OF DIFFERENT METHODS ON THE HOUSTON2013 DATASET \\
Our Methods are based on 10 bands with SVM and CNN\\
\textcolor{blue}{Blue Represents the Best method}, \textcolor{red}{Red Represents the second place model}}
\begin{tabular}{|c||c||c|c|c|c|c||c|c|}
\hline
\hline
\textbf{Class No} & \textbf{CNN-HSI}  & \textbf{CoupledCNNs} & \textbf{EndNet} & \textbf{FusAtNet}&\textbf{Middle-Fusion} &CCL&OurMethod1& \textbf{OurMethod2}\\ 
&\textbf{\cite{yu2017convolutional}}&\cite{hang2020classification}&\cite{hong2020deep}&\cite{mohla2020fusatnet}&\cite{hong2020more}&\cite{jia2023collaborative}&\textbf{SVM}&\textbf{CNN}\\
\hline
\hline
C1  &0.9156&0.8770    & 0.8616   & 0.8150  & 0.8789 &0.8961&  \textcolor{blue}{0.9800}& \textcolor{red}{0.9760} \\
C2 & 0.9253& \textcolor{red}{0.9211}  & 0.8908   & {0.8125}   & 0.9068   &0.9013& 0.8600&\textcolor{blue}{0.9240}  \\
C3  & 0.8869&0.9689   & \textcolor{red}{0.9980}& {0.9588}   & \textcolor{red}{0.9980} &0.9972&0.9500&\textcolor{blue}{1.0000} \\
C4 &0.9410&{0.9774}    & 0.9203   & 0.9202  & 0.9556 &\textcolor{red}{0.9868}  &\textcolor{blue}{1.0000}&0.9770\\
C5 & 0.9913&{0.9920} & {0.9685} & 0.8904 & 0.9919 &\textcolor{red}{0.9958}& 0.9900&\textcolor{blue}{0.9991} \\
C6 &0.8951 &{0.9114}     & 0.8724  & {0.7559}   & 0.8667&0.9203& 0.9200&\textcolor{blue}{0.9860} \\
C7  &0.6912&0.8688    & 0.8013   & 0.7267   & 0.8527  &\textcolor{red}{0.9490}& 0.9400& \textcolor{blue}{0.9930}  \\
C8  & 0.5554&0.7387     & 0.7062   & 0.6543  & 0.7363   &\textcolor{red}{0.8806}& 0.8400&\textcolor{blue}{1.0000} \\
C9 & 0.7275&{0.7547}    & 0.7262  & 0.6152  & 0.7030   &0.8347& \textcolor{red}{0.9400}& \textcolor{blue}{0.9970}  \\
C10 &0.8052 &{0.8270}     & 0.6424 & 0.5933  & 0.7085  & \textcolor{red}{0.9054}&0.7800&\textcolor{blue}{0.9900} \\
C11  & 0.7113&0.9054 & 0.7701  & 0.7211 & 0.8206  &\textcolor{red}{0.9098}& 0.8400 &\textcolor{blue}{0.9820}\\
C12 &0.6406&{0.7934} & 0.6665  & 0.6814   & 0.7760   &0.8365 &\textcolor{red}{0.9100}&\textcolor{blue}{ 0.9770} \\
C13 &0.8045&\textcolor{red}{0.9540} & 0.4680  & 0.7946   & \textcolor{red}{0.9388}  &\textcolor{blue}{0.9663} & 0.8700& 0.9210  \\
C14 &0.9741 &\textcolor{blue}{0.9967} & 0.9737 & 0.9694 &\textcolor{blue}{ 0.9919 }&0.9878&  0.9000&0.9840\\
C15 &0.9989 & {0.9975}   & 0.9832  & 0.9218   & \textcolor{red}{0.9988}  & 0.9955& \textcolor{blue}{1.0000}&0.9980\\ 
\hline
OA &0.8123 &{0.8835}   & 0.8098  & 0.7693 & 0.8576 & {0.9215}&\textcolor{red}{0.9226}& \textcolor{blue}{0.9887} \\
AA & 0.8309&{0.8989}  & 0.8166 & 0.7887   & 0.8758  &0.9306&\textcolor{red}{0.9421}& \textcolor{blue}{0.9912} \\
Kappa & 0.7973 &{0.8741} & 0.7944& 0.7508 & 0.8461  &0.9151 & \textcolor{red}{0.9328}& \textcolor{blue}{0.9878}\\
\hline
\hline
\end{tabular}
\label{uh_fusion_com}
\end{table*}

Our analysis reveals that our fusion models, particularly when evaluated on OA, AA, and Kappa metrics, demonstrate superior performance over the second-place contender, FuseAtNet.

\subsubsection{Fusion comparison based on Trento Data Set}
Table~\ref{tr_fusion_com} illustrates that our SVM fusion model with 10 selected bands has a class C4 in the first place, our CNN fusion model with 10 selected bands has five classes, C1, C2, C4, C5 and C6 in the first place, FuseAtNet has 2 classes in the first dimension, 2 classes in the second dimension. 

\begin{table*}[htp!]
\centering
\caption{CLASSIFICATION PERFORMANCE OF DIFFERENT METHODS ON THE TRENTO DATASET \\
Our Methods are based on 10 bands with SVM and CNN\\
\textcolor{blue}{Blue Represents the Best method}, \textcolor{red}{Red Represents the second place model}}
\begin{tabular}{|c||c||c|c|c|c|c||c|c|}
\hline
\hline
\textbf{Class No} & \textbf{CNN-HSI}  & \textbf{CoupledCNNs} & \textbf{EndNet} & \textbf{FusAtNet}&\textbf{Middle-Fusion} &CCL&OurMethod1& \textbf{OurMethod2}\\ 
&\textbf{\cite{yu2017convolutional}}&\cite{hang2020classification}&\cite{hong2020deep}&\cite{mohla2020fusatnet}&\cite{hong2020more}&\cite{jia2023collaborative}&\textbf{SVM}&\textbf{CNN}\\
\hline
\hline
C1 & 0.9579 & 0.9849 & 0.6035 & \textcolor{red}{0.9709} & 0.9928 &0.9949& 0.9700  & \textcolor{blue}{0.9990}  \\
C2 & 0.8655 & 0.9504 & 0.9468 & \textcolor{red}{0.9519} & 0.9835&0.9760 & 0.9600  & \textcolor{blue}{0.9890} \\
C3 & 0.9718 & 0.9548 & 0.8631 & \textcolor{blue}{0.9049} & 0.9588 & \textcolor{red}{0.9753}&  0.9700  &0.9490  \\
C4 & 0.9843 & 0.9974 & 0.9868& \textcolor{blue}{0.9959} & \textcolor{red}{0.9997} &\textcolor{blue}{1.0000} & \textcolor{blue}{1.0000} & \textcolor{blue}{1.0000}\\
C5 & 0.9678 & \textcolor{red}{0.9991}& 0.8505& 0.9906 & 0.9945 &0.9977 & 0.9900 & \textcolor{blue}{1.0000}\\
C6 & 0.8314 & 0.9306  & 0.8642  & 0.9188   &0.9282& 0.9608 & \textcolor{red}{0.9800} & \textcolor{blue}{0.9930}\\

\hline
OA &  0.9230& 0.9841 & 0.8694 & \textcolor{red}{0.9770} & 0.9873 & 0.9917 & {0.9867}& \textcolor{blue}{0.9969}\\
AA &  0.8601  & 0.9695  & 0.8525 & 0.9555  & 0.9763  & 0.9841 & \textcolor{red}{0.9872}&\textcolor{blue}{0.9942}\\
Kappa & 0.8971 & 0.9788   & 0.8255  & 0.9693   & 0.9831 & \textcolor{red}{0.9890} & 0.9821& \textcolor{blue}{0.9949}\\
\hline
\hline
\end{tabular}
\label{tr_fusion_com}
\end{table*}

Regarding OA, AA and Kappa, our selected 10 bands fused with a 5 layers CNN are at the first place. 

\subsubsection{Fusion Comparison based on MUUFLE Data Set}

MUUFL data contain 11 classes.  Similarly to UH2013 and Trento, we continue to use 10 selected bands to integrate LiDAR data as the input of SVM and 5 layers CNN to conduct experiments. Table~\ref{mf_fusion_com} illustrates that our SVM fusion model with 10 selected bands has a class C1 in the first place and a class C9 in the second place. The CNN fusion model has 1 class C8 in the first place and 4 classes at the second place. 

\begin{table*}[ht]
\centering
\caption{CLASSIFICATION PERFORMANCE OF DIFFERENT METHODS ON THE MUUFL DATASET\\
Our Methods are based on 10 bands with SVM and CNN\\
\textcolor{blue}{Blue Represents the Best method}, \textcolor{red}{Red Represents the second place model}}
\begin{tabular}{|c||c||c|c|c|c|c||c|c|}
\hline
\hline
\textbf{Class No} & \textbf{CNN-HSI}  & \textbf{CoupledCNNs} & \textbf{EndNet} & \textbf{FusAtNet}&\textbf{Middle-Fusion} &CCL&OurMethod1& \textbf{OurMethod2}\\ 
&\textbf{\cite{yu2017convolutional}}&\cite{hang2020classification}&\cite{hong2020deep}&\cite{mohla2020fusatnet}&\cite{hong2020more}&\cite{jia2023collaborative}&\textbf{SVM}&\textbf{CNN}\\
\hline
\hline
C1  & 0.7278&0.8149 & 0.7788  & 0.7349 & 0.7966  &0.8795 & \textcolor{blue}{0.9900} & \textcolor{red}{0.9840}   \\
C2 & \textcolor{blue}{0.8098}&0.6844 & {0.7067} & 0.5400& 0.6424 & 0.7425& 0.6900 & \textcolor{red}{0.7790} \\
C3 & 0.6769&0.5659 & 0.7354 & 0.5176 & 0.6597 & 0.6264& \textcolor{red}{0.8400} &\textcolor{blue}{0.8770} \\
C4  & 0.7126&\textcolor{red}{0.8910}& 0.7539  & {0.6512}  & 0.8062   &\textcolor{blue}{0.9041}& 0.7300 & 0.8590   \\
C5 &0.7599 &0.7861 & 0.8804 & 0.8380 & 0.8331 &{0.8009}&  \textcolor{red}{0.9700}  & \textcolor{blue}{0.9840 }   \\
C6 & 0.9674 &\textcolor{blue}{1.0000}  & \textcolor{red}{0.9923}  & 0.9825 & 0.9982  &\textcolor{blue}{1.0000}& 0.7100  &  0.6880  \\
C7 & \textcolor{blue}{0.9453} & 0.8935 & 0.8938 & 0.7269& \textcolor{red}{0.9246} &0.8125& 0.5100  & 0.7790      \\
C8 &0.7375 &0.8789 & 0.8815 & 0.8539   & 0.8829  &0.8533& \textcolor{blue}{0.9400}  &   \textcolor{red}{0.9330}   \\
C9 & 0.5174&{0.4741} & 0.5478 & 0.3939 & 0.4980  &0.4792& \textcolor{red}{0.6600}  &\textcolor{blue}{ 0.8860}   \\
C10 &\textcolor{blue}{0.7984} &0.6532 & 0.7637 & {0.4618}& 0.5653  &0.6156&  0.1500  & \textcolor{red}{0.7830}  \\
C11 &{0.7361} &0.7625 & 0.7032& {0.7498} & {0.8591} &\textcolor{blue}{0.8655}& 0.8000  &  \textcolor{red}{0.8610} \\
\hline
OA &0.7384 & 0.7743& 0.7915& {0.7076}  & 0.7806 &0.8111& \textcolor{red}{0.8591} & \textcolor{blue}{0.9031}  \\
AA & 0.7626&0.7640  & 0.7852 & 0.6773   & 0.7697  &0.7800& \textcolor{red}{0.8766}  & \textcolor{blue}{0.9026}    \\
Kappa & 0.6748&0.7133  & 0.7366  & {0.6337}  & 0.7217   &0.7562&\textcolor{red}{0.8189}  &  \textcolor{blue}{0.8739}  \\
\hline
\hline
\end{tabular}
\label{mf_fusion_com}
\end{table*}
As for OA, AA, and Kappa of the classification performance based on MUUFL data set,  although our method OA and  are at the second place, the AA is at the first place. 

Three different datasets have validated our proposed method, showing that our method consistently outperforms traditional and contemporary band selection techniques. By utilising both spatial and spectral information, our method adeptly navigates the complex data landscape, ensuring that the selected bands are not only relevant, but also conducive to higher classification accuracy.

The experiments underscore the importance of feature separability in achieving high classification performance. Future research could investigate the development of more advanced metrics for evaluating feature separability, beyond traditional distance criteria. Such metrics could account for the complex interplay between bands and the multidimensional nature of hyperspectral data, offering more nuanced assessments of band importance and redundancy.

\section{Conclusions} \label{sec:conclusions}
In our study, we propose a novel unsupervised approach to band selection in hyperspectral image (HSI) processing. Our method, distinct from conventional techniques, incorporates an attention module within an Autoencoder architecture, adeptly harnessing both the spatial and spectral attributes of HSIs alongside the spatial features of LiDAR data. This strategy effectively counters the issue of hyperspectral redundancy by selecting a concise yet highly informative set of bands, thus optimising the efficiency of subsequent analytical tasks and reducing computational overhead.

At the heart of our proposed solution is the generation of a focus-fused feature mask synthesising input from both HSI and LiDAR data, which undergoes processing via a convolutional Autoencoder. This innovative step ensures a thorough encapsulation of spatial and spectral relevance within the imagery. To facilitate the selection of informative bands, we devise a unique combination distance metric, merging attention and dissimilarity distances, which, alongside a clustering algorithm, aids in the meticulous selection of band subsets.

Our evaluation spanned three comprehensive datasets: Houston 2013, Trento and MUUFL, where our method demonstrated superior performance against existing unsupervised hyperspectral band selection techniques in classification accuracy. This evaluation, carried out through the implementation and evaluation of each compared method according to its foundational specifications, ensured a rigorous and equitable comparison.

The core of our proposed method lies in the generation of a focus-fused feature mask derived from both HSI and LiDAR inputs, which is then processed through a convolutional Autoencoder. This novel approach allows for a comprehensive capture of both spatial and spectral significance within the hyperspectral imagery. To efficiently select informative bands, we introduced a custom combination distance comprising the attention distance and the dissimilarity distance, which serves as the criterion for the selection of small-band sets aided by a clustering algorithm.

Conclusively, our approach constitutes a significant contribution to the field of hyperspectral imaging analysis, particularly beneficial in situations marked by limited labeled data and the integration of auxiliary data types. Although our results are encouraging, further exploration into incorporating diverse auxiliary data, refining the attention mechanism, and broadening the application spectrum of our methodology in remote sensing tasks remains a promising direction for future research. Our method's adaptability and efficiency signify a substantial stride towards enhancing hyperspectral image processing capabilities, heralding advancements in this rapidly progressing area.



\bibliographystyle{IEEEtran}

\bibliography{IEEEabrv, BS_Autoencoder_Fusion_LiDAR_HSI. bib}

\vfill

\end{document}